# Niching an Archive-based Gaussian Estimation of Distribution Algorithm via Adaptive Clustering

Yongsheng Liang
Autocontrol Institute, Xi'an Jiaotong University
Xi'an, Shaanxi, 710049, P.R. China
liangyongsheng@stu.xjtu.edu.cn

Zhigang Ren
Autocontrol Institute, Xi'an Jiaotong University
Xi'an, Shaanxi, 710049, P.R. China
renzg@mail.xjtu.edu.cn

Bei Pang
Autocontrol Institute, Xi'an Jiaotong University
Xi'an, Shaanxi, 710049, P.R. China
beibei@stu.xjtu.edu.cn

An Chen
Autocontrol Institute, Xi'an Jiaotong University
Xi'an, Shaanxi, 710049, P.R. China
chenan123@stu.xjtu.edu.cn

## ABSTRACT

As a model-based evolutionary algorithm, estimation of distribution algorithm (EDA) possesses unique characteristics and has been widely applied to global optimization. However, traditional Gaussian EDA (GEDA) may suffer from premature convergence and has a high risk of falling into local optimum when dealing with multimodal problem. In this paper, we first attempts to improve the performance of GEDA by utilizing historical solutions and develops a novel archive-based EDA variant. The use of historical solutions not only enhances the search efficiency of EDA to a large extent, but also significantly reduces the population size so that a faster convergence could be achieved. Then, the archive-based EDA is further integrated with a novel adaptive clustering strategy for solving multimodal optimization problems. Taking the advantage of the clustering strategy in locating different promising areas and the powerful exploitation ability of the archive-based EDA, the resultant algorithm is endowed with strong capability in finding multiple optima. To verify the efficiency of the proposed algorithm, we tested it on a set of well-known niching benchmark problems and compared it with several state-of-the-art niching algorithms. The experimental results indicate that the proposed algorithm is competitive.

## 1 INTRODUCTION

Estimation of distribution algorithm (EDA) [1-3] is a special branch of evolutionary algorithm (EA). Different from other EAs, the new solutions of EDA are generated by sampling from a probability distribution. The probability distribution is generally estimated from some high-quality solutions selected in the current generation. It is hoped that the estimated probability distribution could capture the structural characteristics of the problem, thus effectively guiding the optimization process. Since it came into being, EDA has attracted increasing research effort and achieved great success in both combinatorial and continuous domains [4]. In this paper, EDAs for continuous domain are studied.

Continuous EDA usually adopts Gaussian model as the basic probability distribution model. According to the variable dependencies, Gaussian EDA (GEDA) can be classified into three kinds, including univariate GEDA [1], bivariate GEDA [2] and multivariate GEDA [3], among these three kinds of GEDA, multivariate GEDA shows competitive performance on most kinds of problems since it can describe the variable dependencies well, but it usually requires a large population to build a feasible multivariate model. A representative algorithm that employs multivariate model is estimation of multivariate normal density algorithm (EMNA$_g$) [3].

Traditional GEDAs possess powerful exploitation ability and perform well on some simple unimodal problems, but they may suffer from premature convergence in relatively complicated cases [5, 6]. To solve that, many techniques were successfully developed such as adaptive variance scaling (AVS) [5] and anticipated mean shift (AMS) [6]. The combination of AVS and AMS leads to an efficient EDA variant known as AMaLGaM [6]. Nevertheless, GEDAs still have difficulty in dealing with multimodal problems. The main reason lies in that the structure of multimodal problem cannot be well represented by a unimodal Gaussian distribution [7], so GEDA is usually limited in only one promising region and has a high risk of ending up in a local optimum. Multimodal problems naturally call for more complicated models, Gaussian mixture model (GMM) based EDAs have been developed to remedy this disadvantage. A GMM is a weighted sum of single Gaussian models, where each of these single Gaussian model governs a group of solutions. However, the computational cost to learn a precise GMM is usually high [8] and it is non-trivial to set the number of mixture models in advance, especially for black box problems [9].

To improve the performance of EA on multimodal problems, many alternative techniques for locating multiple optima have been developed, commonly referred to as niching methods [10]. Some representative niching methods include clearing [8], crowding [11], speciation [11] and clustering [7, 9, 12]. Based on these methods, many niching EDAs were successfully suggested. Dong and Yao [8] proposed a NichingEDA which employs the clearing strategy to maintain the diversity of subpopulations so that they could explore different promising regions. Yang et al. [13] proposed a novel maintaining and processing multiple sub-models technique to enhance the performance of EDA on multimodal problems. Yang et al. [11] developed tow multimodal EDAs (MEDAs) based on crowding and speciation, respectively. They further enhanced the two MEDAs with dynamic cluster-sizing strategy and local search scheme for relieving the parameter sensitivity and improving the solution accuracy, the resultant algorithms were named as LMCEDA and LMSEDA. In addition, clustering strategies are also widely used to do niching. References [7] and [12] utilized adaptive rival penalized competitive learning clustering and affinity propagation clustering, respectively, to adaptively partition the population of EDA. Maree et al. [9] presented a hierarchical Gaussian mixture learning method (HGML) that determines the number of niches automatically based on hierarchical clustering. Then they integrated HGML with AMaLGaM and developed an efficient niching algorithm called clustered AMaLGaM (CAMaLGaM).

In summary, the performance of EDA on multimodal problems has been significantly improved by using niching methods, but there are still some shortcomings. First, many niching methods are sensitive to the parameter setting. Clearing and speciation are both radius-based methods, they rely on a predefined niche radius to check whether two solutions belong to the some niche. Crowding method is sensitive to the crowding size. Second, most of these niching EDAs adopt the basic EDA to search the optima after locating promising areas, which may reduce their overall efficiency since traditional GEDA usually suffers from premature convergence. MEDAs and CAMaLGaM demonstrate competitive performance in locating multiple optima, but they are accompanied by more complex algorithmic framework and parameters. Besides, except for MEDAs and CAMaLGaM, the other algorithms aforementioned are all designed for single global optimization.

To alleviate these deficiencies, this paper first proposes a novel archive-based EDA variant named EDA$^2$, then EDA$^2$ is further incorporated into an adaptive clustering strategy for solving multimodal problems. Instead of only utilizing solutions in current generation, EDA$^2$ preserves some high-quality historical solutions into an archive and takes advantage of these historical solutions to assist estimating the Gaussian model. By this means, the evolution direction information is naturally integrated into the estimated model which in turn can efficiently improve the search ability of EDA$^2$. The use of historical solutions also reduces the population size of EDA$^2$ so that a faster convergence could be achieved. Since the performance of EDA is enhanced by exploiting Evolution Direction information hidden in the Archive, we named this algorithm EDA$^2$. By combining EDA$^2$ with an adaptive clustering strategy that could adaptively detect different promising regions based on the fitness value and relative distance of solutions, the resultant algorithm, referred to as C-EDA$^2$, shows appealing performance in dealing with multimodal problems. To verify its efficiency, extensive experiments were executed on the CEC'2013 niching benchmark problems. Experimental results demonstrate that C-EDA$^2$ is competitive and has some potential to be improved.

The remainder of this paper is organized as follows. Section 2 briefly reviews the estimation of distribution algorithm. Section 3 presents EDA$^2$, the clustering strategy and the resultant C-EDA$^2$. Experimental results are reported in Section 4，and conclusions are drawn in Section 5.

## 2 ESTIMATION OF DISTRIBUTION ALGORITHM

As a model-based EA, EDA assumes that good solutions approximately obey a certain probability distribution over the solution space. It tries to learn this distribution and generate new solutions according to the learning results [1-3]. The general framework of EDA is outlined in Algorithm 1.

| **Algorithm 1**: General framework of EDA |
| --- |
| 1. Initialize parameters, set $t = 0$, and generate the initial population $P^t$; |
| 2. Evaluate $P^t$ and update the best solution $b^t$ obtained so far; |
| 3. Output $b^t$ if the stopping criterion is met; |
| 4. Select promising solutions $S^t$ from $P^t$; |
| 5. Build a new probability model $G^{t+1}$ based on $S^t$, update $t \leftarrow t + 1$; |
| 6. Generate a new population $P^t$ by sampling from $G^t$ and goto step 2. |

Continuous EDAs generally employ Gaussian model as the basic probability distribution model. The Gaussian probability density function for an $n$-dimensional random vector $\bm{x}$ can be parameterized by its mean $\bm{\mu}$ and covariance matrix $\bm{C}$ as follows:

$$G_{(\bm{\mu},\bm{C})}(\bm{x}) = \frac{(2\pi)^{-n/2}}{(\det \bm{C})^{1/2}} \exp(-(\bm{x}-\bm{\mu})^{\mathrm{T}} (\bm{C})^{-1}(\bm{x}-\bm{\mu})/2) . \quad (1)$$

$\bm{\mu}$ and $\bm{C}$ for the next generation are generally estimated according to the maximum likelihood estimation method based on the solutions selected from the current population:

$$\bar{\bm{\mu}}^{t+1} = \frac{1}{|\bm{S}^t|} \sum_{i=1}^{|\bm{S}^t|} \bm{S}_i^t , \quad (2)$$

$$\bar{\bm{C}}^{t+1} = \frac{1}{|\bm{S}^t|} \sum_{t=1}^{|\bm{S}^t|} (\bm{S}_i^t - \bar{\bm{\mu}}^{t+1})(\bm{S}_i^t - \bar{\bm{\mu}}^{t+1})^{\mathrm{T}} , \quad (3)$$

where $\bm{S}^t$ denotes the set of solutions selected from the current population. The Gaussian model estimated by Eqs. (2) and (3)

takes the dependencies between all pairs of variables into account. It could ensure rotation-invariance and is capable of capturing some complex structural characteristics of the solution space [6], hence is widely used.

However, the covariance matrix of this model totally have $0.5n^2+0.5n$ free parameters. To make a proper estimation, the number of required samples should be much larger than $n$. For the traditional GEDA, all the samples are selected from the current population. This is the main reason why it requires a much larger population size. If a limited amount of computational resource is available, it can only evolve a few generations. Especially for multimodal optimization where multiple promising regions are of interest, the computational resource for each niche would be further reduced. This will greatly deteriorate the performance of GEDA. Therefore, it is of great significance to study new estimation method that could reduce the population size of GEDA and thus improve its search efficiency.

Furthermore, from the perspective of optimization, the aim of learning Gaussian model in EDA is not to rigidly describe the distribution of high-quality solutions in the current population, but to predict the distribution of new promising solutions thus facilitating the algorithm finding them in subsequent generations. Based on this point of view, it is reasonable to exploit historical solutions rather than just the current population for model estimation since they could reveal the evolution information of good solutions. According to this idea, this paper proposes a novel archive-based EDA variant named EDA² which would be shown to be simple and efficient. Then, EDA² is further incorporated into an adaptive clustering strategy for solving multimodal problems.

## 3 DESCRIPTION OF C-EDA²

This section first introduces the two components of C-EDA², i.e. EDA² and the adaptive clustering strategy, then presents the full procedure of C-EDA² in detail.

### 3.1 Archive-based EDA: EDA²

During the optimization process, EDA generally builds the probability model based on the current population. Historical solutions produced in previous generations are usually abandoned, although they may contain some meaningful information. To exploit such information, EDA² maintains an archive to store a certain number of historical high-quality solutions and estimates the covariance matrix of Gaussian model by using these solutions as well as the ones selected from current generation. For each generation, the archive $A^t$ is defined as follows:

$$A^t = S^{t-1} \cup S^{t-2} \cup ... \cup S^{t-l}, \qquad (4)$$

where $S^{t-i}$ denotes the set of solutions selected at the $(t-i)$th generation, $l$ is a nonnegative integer and denotes the length of the archive. This means that EDA² preserves the solutions selected at the last $l$ generations into its archive. The archive would be empty if we set $l$ equal to zero.

Once $A^t$ is determined, EDA² estimates its covariance matrix as follows:

$$\overline{C}^{t+1} = \frac{1}{|H^t|}\sum_{i=1}^{|H^t|}(H_i^t - \overline{\mu}^{t+1})(H_i^t - \overline{\mu}^{t+1})^T, \ H^t = A^t \cup S^t, \qquad (5)$$

where the new mean $\overline{\mu}^{t+1}$ is still estimated according to Eq. (2) which implies it only depends on $S^t$.

It is known that $\overline{\mu}$ and $\overline{C}$ determine the search characteristics of GEDA, which can be geometrically described by a probability density ellipsoid (PDE) in the hyperspace [6]. A schematic is shown in Fig. 1 to demonstrate the effect of the new estimation method in EDA². PDE-0, PDE-1 and PDE-$l$ in Fig. 1 schematically represent three new PDEs estimated by Eq. (5) with archives of length 0, 1 and $l$, respectively, where PDE-0 could represent the PDE of traditional multivariate GEDA. It can be seen that PDE estimated by the solutions selected at several consecutive generations according to Eq. (5) will be approximately elongated along the movement direction of corresponding means, i.e., along the evolution direction of EDA². And the greater the archive length is, the further PDE will be elongated.

The new estimation method naturally integrates the evolution direction information into the estimated covariance matrix, which endows EDA² with better search direction and greater search scope. With a proper value for archive length $l$, the search efficiency of EDA² is expected to be greatly enhanced. Moreover, the new estimation method can greatly reduce the population size of EDA² since solutions selected from ($l$+1) generations rather than only the current generation are used to estimate the covariance matrix.

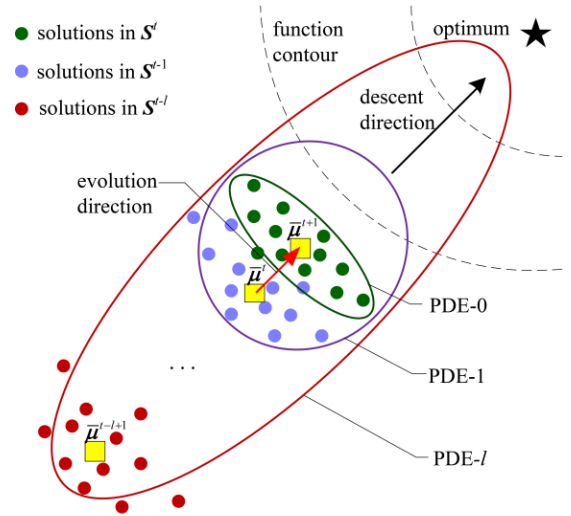

**Figure 1: Schematic of PDEs estimated with different archive lengths in EDA².**

Algorithm 2 presents the detailed steps of EDA², where three points should be noted. First, EDA² initializes archive to be empty (step 2) and constantly adds the solutions selected at each generation to archive until the archive length reaches the specified value (steps 8-9). If that is the case, EDA² updates archive by replacing the oldest set of selected solutions with the latest one (steps 10-11). Second, EDA² employs the commonly used truncation selection rule to select solutions from the current population (step 5). Finally, EDA² also takes an elite strategy which maintains the best solution at the current generation to the next generation. Therefore, it just generates $p-1$ new solutions for the population of size $p$ (steps 13-14).

---

**Algorithm 2:** Procedure of EDA²

1. Initialize parameters, including population size $p$, selection ratio $\tau$, and archive length $l$;
2. Set $t = 0$, $i = 0$, and $A^t = \emptyset$, and randomly generate the initial population $P^t$;
3. Evaluate $P^t$ and update the best solution $b^t$ obtained so far;
4. Output $b^t$ if the stopping criterion is met;
5. Select the best $\lfloor \tau p \rfloor$ solutions from $P^t$ and store them into $S^t$;
6. Estimate the mean $\bar{\mu}^{t+1}$ with $S^t$ according to (2);
7. Estimate covariance matrix $\bar{C}^{t+1}$ with $S^t$ and $A^t$ according to (5);
8. **If** $i < l$ **then**
9.     set $A^{t+1} = A^t \cup S^t$ and set $i \leftarrow i + 1$;
10. **Else**
11.     $A^{t+1} = A^t \cup S^t \setminus S^{t-l}$;
12. Set $t \leftarrow t + 1$, build a probability model $G^t$ based on $\bar{\mu}^t$ and $\bar{C}^t$ ;
13. Generate $p - 1$ new solutions by sampling from $G^t$ and store them into $M^t$;
14. Set $P^t = M^t \cup b^{t-1}$ and goto step 3.

---

## 3.2 Adaptive Clustering Strategy

Decision space and target space (DS-TS) information based clustering is adopted in C-EDA² to adaptively capture different promising regions without using any prior knowledge. DS-TS clustering is inspired by the thought of a fast clustering strategy proposed in [14]. The fast clustering strategy is based on the assumption that cluster centers are characterized by relatively higher density and larger distance, it is mainly used for image clustering. Considering the characteristics of optimization problem, the basic idea of DS-TS clustering consists in that cluster centers are solutions with better fitness value and farther relative distance.

Supposing there are $m$ solutions that need to be clustered, the procedure of DS-TS clustering is described by the following steps. Without loss of generality, maximization problem is assumed in here.

1) Sorting solutions in ascending order according to their fitness values;

2) Computing the relative distance of each solution;

For the $i$ th solution, its relative distance $\delta_i$ is defined as the minimum distance between it and any other solution with better fitness value:

$$\delta_i = \min_{j:\, f(j) > f(i)} (d_{ij}),\qquad(6)$$

where $f(i)$ is the fitness value of the $i$ th solution, $d_{ij}$ is the Euclidean distance between the $i$ th and $j$ th solution. Particularly, for the best solution $x_m$, since there is no better solution than it, we directly define its relative distance $\delta_m$ as:

$$\delta_m = \max_{1 \leq i < m} (\delta_i).\qquad(7)$$

3) Computing the distance threshold and choosing cluster centers;

The distance threshold $\delta_{th}$ is defined by Eq. (8), where α is the threshold factor. Then solutions who satisfy Eq. (9) would be chosen as cluster centers.

$$\begin{cases} \delta_{max} = \max_{1 \leq i \leq m} (\delta_i) \\ \delta_{min} = \min_{1 \leq i \leq m} (\delta_i) \\ \delta_{th} = \alpha \cdot (\delta_{max} - \delta_{min}) \end{cases},\qquad(8)$$

$$centers = \{x_i | \delta_i > \delta_{th}\}.\qquad(9)$$

4) Assigning cluster members;

After determining the cluster centers, each remaining solution is assigned to the same cluster as its nearest neighbour of better fitness value.

Fig. 2 shows a simple example to illustrate the effect of DS-TS clustering in one-dimensional space. It can be seen from Fig. 2 that points 2 and 6 have better fitness values and farther relative distances, they would be chosen as cluster centers. Then, points 1 and 3 are assigned to the cluster of point 2. Point 4 belongs to the same cluster with point 5, and they are both assigned to the cluster of point 6. DS-TS clustering is very simple and fast, it has been shown to perform well in the former work [15]. The only parameter in DS-TS clustering, i.e. the threshold factor α, should be generally set within (0, 1).

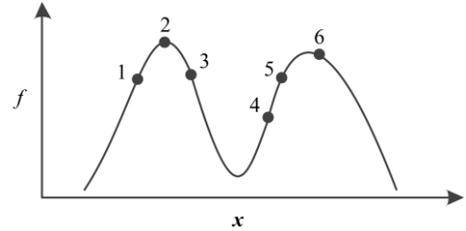

**Figure 2: Example of DS-TS clustering.**

## 3.3 Procedure of C-EDA²

The design of C-EDA² is to cluster a group of selected solutions into clusters by DS-TS clustering strategy, then EDA² is utilized to evolve these clusters independently and find their optima. Algorithm 3 presents the procedure of C-EDA², where several points should be noted. First, the truncation selection ratio $\tau$ in step 3 is set to the same with that in EDA². Second, we use EDA² with the same parameter setting to evolve different clusters independently (step 6). Since the number of solutions in different clusters may be different, in the first generation of EDA², different numbers of solutions would be used to estimate the Gaussian model, while the rest steps of EDA² are the same with that in Algorithm 2. Besides, EDA² would be stopped if one of the following termination criteria is met: 1) the improvement of the median of population values is smaller than the defined accuracy level in the last 5 generations; 2) the maximum number of function evaluations (MaxFEs) is reached.

---

**Algorithm 3**: Procedure of C-EDA$^2$

1. Set the initial solution number *N*, selection ratio $\tau$, initialize EDA$^2$ and DS-TS clustering, set *Output*= ∅;
2. **while** the MaxFEs is not reached **do**
3.    Randomly and uniformly initialize *N* solutions and select the best $\lfloor \tau N \rfloor$ solutions of them;
4.    Use DS-TS clustering strategy to divide the selected solutions into different clusters;
5.    **for** *k* = 1 to *the number of clusters* **do**
6.       Use EDA$^2$ to evolve the solutions in the *k* th cluster until the termination criteria is met;
7.       Store the best solution obtained into *Output*;
8.    **end**
9. **end**
10. Output the solutions in *Output*.

---

## 4 EXPERIMENTAL STUDY

This section aims to study the influence of parameters, investigate the effectiveness of EDA$^2$ and DS-TS clustering strategy, and synthetically evaluate the performance of C-EDA$^2$ by testing it on the 20 niching benchmark problems of CEC'2013 special session on multimodal optimization.

### 4.1 Influence of Parameters in EDA$^2$

There are only three parameters in EDA$^2$, including selection ratio $\tau$, population size *p* and archive length *l*. The selection ratio $\tau$ is conventionally set as $\tau = 0.35$, this section mainly focus on studying the influence of *p* and *l*. To achieve that, the performance of EDA$^2$ is systematically tested on a set of optimization problems in [16] with different values of *p* and *l*. For brevity, here we only show the experiments on two typical test functions in 20 dimensions, including the high conditioned elliptic function and Rosenbrock's function. High conditioned elliptic function is a unimodal function and Rosenbrock's function is a multimodal function, they are both shifted and rotated to increase the solving difficulty. Details of this two functions can be found in [16].

In this experiment, the maximum number of function evaluations (FEs) in a single run is set to 200,000 and 25 independent runs are conducted on each function. The performance of EDA$^2$ is evaluated according to the function error value (FEV) of the obtained best solution, i.e. the difference between its fitness value and that of the global optimum. Note that the FEV will be reported as zero if it is smaller than $10^{-8}$.

The values of *p* and *l* consider in this experiment include $p \in$ {50, 80, 110, 140, 170, 200} and $l \in$ {5, 10, 15, 20, 25, 30}. Fig. 3 shows the results obtained by EDA$^2$ with different combinations of *p* and *l*. It can be found that EDA$^2$ performs surprisingly well on the two functions when *p* and *l* are located in a valley-like region, which is very meaningful since it reveals that *p* and *l* can complement each other. A small value for one parameter coupled with a large value for the other or moderate values for both of them could always achieve satisfying performance.

To further investigate the influence of *p* and *l* on the evolution process of EDA$^2$, Fig. 4 presents the evolution curves of the average FEVs obtained by three different EDA$^2$s whose parameter settings are:

1) EDA$^2$-1: *p* = 80, *l* = 10;
2) EDA$^2$-2: *p* = 80, *l* = 15;
3) EDA$^2$-3: *p* = 140, *l* = 10.

It can be seen from Fig. 4 that EDA$^2$-1 demonstrates the fastest convergence speed on the two functions and achieves superior result on the first unimodal function, but it is more likely to fall into local optimum of the second function. Compared to EDA$^2$-1, EDA$^2$-2 and EDA$^2$-3 have larger value of *l* and *p*, respectively. It can be observed from Fig. 4 that EDA$^2$-2 and EDA$^2$-3 indicate stronger exploration ability and could find the optimal solution of the Rosenbrock's function, but their convergence speed on the first function is slowed down. For reference, Fig. 4 also presents the evolution curve of a traditional multivariate GEDA, i.e. EMNA$_g$ [3], whose population size is set to 500. It is obvious that EMNA$_g$ converges prematurely on both functions.

In summary, EDA$^2$ is very efficient, which demonstrates the effectiveness of exploiting evolution direction information with archive. Moreover, EDA$^2$ is also robust to its parameters, the settings of *p* and *l* usually have many options. Relatively small values of *p* and *l* endow EDA$^2$ with faster convergence speed, while larger values of *p* and *l* are beneficial to improve the exploration ability. In addition, compared to the traditional multivariate GEDA, the population size of EDA$^2$ is significantly reduced.

### 4.2 Effectiveness of DS-TS Clustering

In order to illustrate the performance of DS-TS clustering, its clustering process on a multimodal function is presented in this section. Six-hump camel back function from the CEC'2013 niching benchmark suite [17] is adopted, which is a two dimensional maximization function with two global optima and two local optima. We uniformly initialize 1000 solutions in the solution space and select the top 35% solutions after evaluating them. Then DS-TS clustering strategy is used to cluster the selected solutions. The threshold factor α of DS-TS clustering is set to 0.8.

Fig. 5 shows the fitness value and relative distance of selected solutions obtained by DS-TS clustering, it is clearly that there are four striking solutions that have relatively better fitness values and farther relative distances. After calculating the distance threshold, these four solutions are chosen as cluster centers and four different clusters could be obtained. Fig. 6 presents the landscape of Six-hump camel back function and the final clustering result, where points and stars with four different colours on the surface represent the cluster members and cluster centers of different clusters. It can be seen from Fig. 6 that DS-TS clustering strategy correctly divides the selected solutions into four clusters, and each cluster covers a promising region of the solution space.

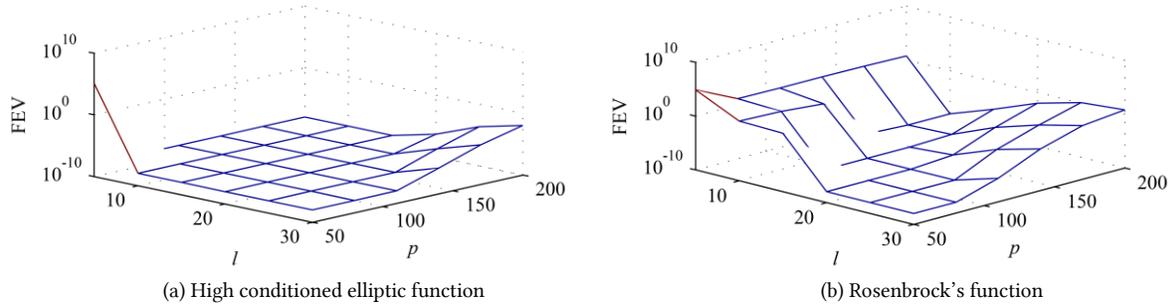

(a) High conditioned elliptic function    (b) Rosenbrock's function

**Figure 3: Average FEVs obtained by EDA² with different combinations of $p$ and $l$.**

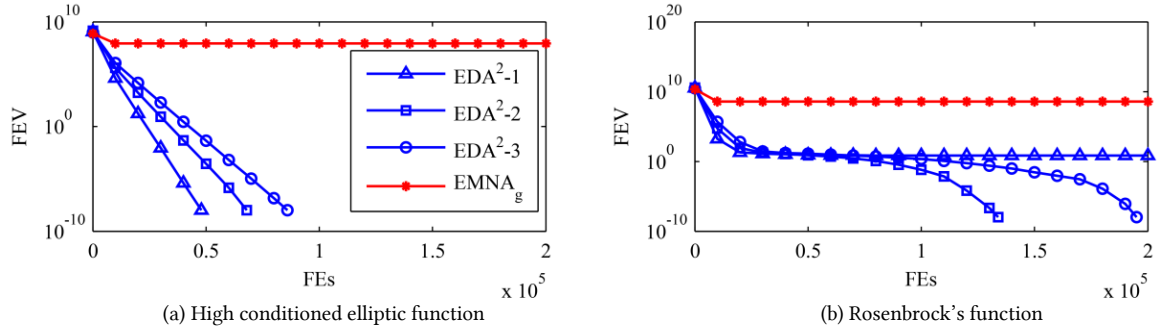

(a) High conditioned elliptic function    (b) Rosenbrock's function

**Figure 4: Evolution of FEVs derived from three EDA²s and EMNA$_g$.**

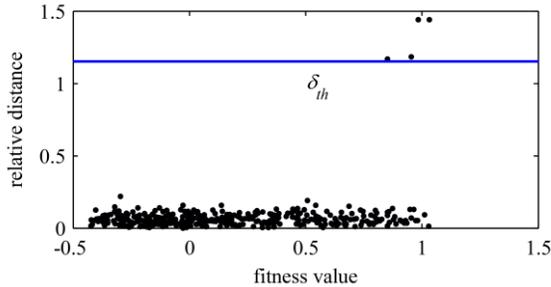

**Figure 5: Fitness values and relative distances of selected solutions.**

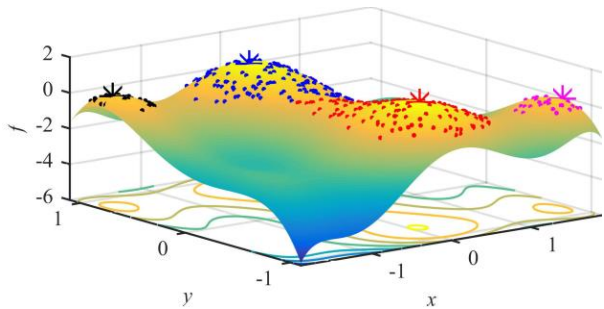

**Figure 6: Landscape of Six-hump camel back function and the clustering result.**

Besides, it can be also observed from Fig. 5 that DS-TS clustering could obtain stable result with a wide range of $\alpha$. Relatively smaller value of $\alpha$ may result in more clusters, while larger value of $\alpha$ can filter out some local regions that are less attractive.

### 4.3 Performance of C-EDA²

In this section, C-EDA² is tested on the CEC'2013 multimodal function set and compared with several state-of-the-art niching algorithms for multimodal optimization. CEC'2013 multimodal function set contains 20 multimodal functions, their main characteristics are summarized in Table 1, including problem dimension $D$, number of global optima (*No. of GO*) and the allowed maximum number of function evaluations (MaxFEs). To evaluate the performance of C-EDA², the peak ratio (PR) is used, which is defined as the percentage of the number of the global optima found out of the total number of global optima averaged over multiple runs. A peak, i.e. a global optimum, is correctly detected if it is within an accuracy level $\varepsilon$ of the true optimum.

The initial solution number $N$ in C-EDA² is set as $N=1000+10D^2$ to improve the exploration ability on high dimensional problems like $f_{20}$. The selection ratio $\tau$ is set to 0.35, and the threshold factor $\alpha$ of DS-TS clustering is set to 0.8. Since the promising solution regions have been located by clustering, we suggest choosing relatively small values of $p$ and $l$ for EDA² to enhance its exploitation ability so that it could find the optimum of each cluster using fewer FEs. For this set of test functions, $p$ and $l$ of EDA² are set as $p = 4(D+1)$ and $l = 5$.

**Table 1: Main characteristics of the 20 functions of CEC'2013 multimodal function set.**

| Fun. | D | No. of GO | MaxFEs |
|---|---|---|---|
| $f_1$ | 1 | 2 | 5.0E+04 |
| $f_2$ | 1 | 5 | 5.0E+04 |
| $f_3$ | 1 | 1 | 5.0E+04 |
| $f_4$ | 2 | 4 | 5.0E+04 |
| $f_5$ | 2 | 2 | 5.0E+04 |
| $f_6$ | 2 | 18 | 2.0E+05 |
| $f_7$ | 2 | 36 | 2.0E+05 |
| $f_8$ | 3 | 81 | 4.0E+05 |
| $f_9$ | 3 | 216 | 4.0E+05 |
| $f_{10}$ | 2 | 12 | 2.0E+05 |
| $f_{11}$ | 2 | 6 | 2.0E+05 |
| $f_{12}$ | 2 | 8 | 2.0E+05 |
| $f_{13}$ | 2 | 6 | 2.0E+05 |
| $f_{14}$ | 3 | 6 | 4.0E+05 |
| $f_{15}$ | 3 | 8 | 4.0E+05 |
| $f_{16}$ | 5 | 6 | 4.0E+05 |
| $f_{17}$ | 5 | 8 | 4.0E+05 |
| $f_{18}$ | 10 | 6 | 4.0E+05 |
| $f_{19}$ | 10 | 8 | 4.0E+05 |
| $f_{20}$ | 20 | 8 | 4.0E+05 |

**Table 2: Average peak ratio obtained by C-EDA$^2$ over 50 independent runs on the 20 functions of CEC'2013 multimodal function set at five different accuracy levels.**

| Fun. | $\varepsilon=10^{-1}$ | $\varepsilon=10^{-2}$ | $\varepsilon=10^{-3}$ | $\varepsilon=10^{-4}$ | $\varepsilon=10^{-5}$ |
|---|---|---|---|---|---|
| $f_1$ | 1.000 | 1.000 | 1.000 | 1.000 | 1.000 |
| $f_2$ | 1.000 | 1.000 | 1.000 | 1.000 | 1.000 |
| $f_3$ | 1.000 | 1.000 | 1.000 | 1.000 | 1.000 |
| $f_4$ | 1.000 | 1.000 | 1.000 | 1.000 | 1.000 |
| $f_5$ | 1.000 | 1.000 | 1.000 | 1.000 | 1.000 |
| $f_6$ | 1.000 | 1.000 | 1.000 | 1.000 | 1.000 |
| $f_7$ | 0.793 | 0.741 | 0.729 | 0.711 | 0.705 |
| $f_8$ | 0.878 | 0.856 | 0.844 | 0.839 | 0.822 |
| $f_9$ | 0.644 | 0.330 | 0.309 | 0.300 | 0.292 |
| $f_{10}$ | 1.000 | 1.000 | 1.000 | 1.000 | 1.000 |
| $f_{11}$ | 0.683 | 0.667 | 0.667 | 0.667 | 0.667 |
| $f_{12}$ | 0.670 | 0.667 | 0.667 | 0.663 | 0.653 |
| $f_{13}$ | 0.673 | 0.667 | 0.667 | 0.667 | 0.667 |
| $f_{14}$ | 0.670 | 0.667 | 0.667 | 0.667 | 0.667 |
| $f_{15}$ | 0.770 | 0.745 | 0.743 | 0.740 | 0.740 |
| $f_{16}$ | 1.000 | 0.667 | 0.667 | 0.667 | 0.667 |
| $f_{17}$ | 1.000 | 0.698 | 0.690 | 0.660 | 0.648 |
| $f_{18}$ | 1.000 | 0.667 | 0.667 | 0.667 | 0.667 |
| $f_{19}$ | 1.000 | 0.500 | 0.500 | 0.500 | 0.500 |
| $f_{20}$ | 0.798 | 0.250 | 0.250 | 0.248 | 0.243 |
| **Average** | **0.879** | **0.756** | **0.753** | **0.750** | **0.747** |

Table 2 reports the optimization results of C-EDA$^2$ on CEC'2013 multimodal function set at five different accuracy levels, i.e., $\varepsilon = 10^{-1}$, $\varepsilon = 10^{-2}$, $\varepsilon = 10^{-3}$, $\varepsilon = 10^{-4}$ and $\varepsilon = 10^{-5}$. All experiments are carried out for 50 independent runs. From Table 2, the following observations can be made:

1) C-EDA$^2$ successfully finds all the global optima of $f_1$-$f_6$ and $f_{10}$ at all accuracy levels, it could also locate all the global optima of $f_{16}$-$f_{19}$ at accuracy level $\varepsilon = 10^{-1}$. Both $f_7$ and $f_8$ contain dozens of global optima, nevertheless, C-EDA$^2$ can find most of them at all accuracy levels. With respect to $f_{11}$-$f_{15}$, C-EDA$^2$ achieves almost the same performance at five accuracy levels. However, the results on $f_9$ and $f_{20}$ are relatively undesirable that C-EDA$^2$ misses most of the global optima when the accuracy level varies from $10^{-2}$ to $10^{-5}$. The reason for the former lies in that $f_9$ has an ocean of global optima that it is hard to locate them all, while the reason for the latter mainly consists in that the solution space is too huge to be fully detected with the increase of problem dimension.

2) From the perspective of accuracy level, C-EDA$^2$ obtains fine results on most functions at accuracy level $\varepsilon = 10^{-1}$, and the average PR on the 20 functions reaches 0.879. While the performance of C-EDA$^2$ degenerates rapidly when the accuracy level changes form $10^{-1}$ to $10^{-2}$, especially on $f_9$ and $f_{20}$. Such phenomenon is common for many multimodal algorithms [11]. As for the other accuracy levels, C-EDA$^2$ shows very similar performance on all the functions and the average PR only decreases a little when $\varepsilon$ varies from $10^{-2}$ to $10^{-5}$, which demonstrates the powerful exploitation ability of EDA$^2$.

To further evaluate the efficiency of C-EDA$^2$, we compared it with LMCEDA [11], LMSEDA [11] and RS-CMSA [18]. LMCEDA and LMSEDA are two well-established multimodal EDAs, which have been briefly introduced in Section 1. RS-CMSA combines the famous covariance matrix self-adaptation evolution strategy [19] with a novel niching technique based on repelling subpopulations, it is the winner of the GECCO'2017 competition on niching methods for multimodal optimization. For saving space, we only compare their results at accuracy level $\varepsilon = 10^{-4}$. Table 3 summarizes the average PRs obtained by the four algorithms, where the results of LMCEDA and LMSEDA are directly taken from their original paper, and the results of RS-CMSA are taken from [20].

From Table 3, the following comments can be made:

1) C-EDA$^2$ achieves better performance than LMCEDA and LMSEDA on most functions. Concretely, C-EDA$^2$ obtains better, same or worse results than LMCEDA on 8, 10 and 2 functions, and the corresponding numbers for LMSEDA are 8, 9 and 3. So C-EDA$^2$ has an edge over the two algorithms.

2) Compared to RS-CMSA, C-EDA$^2$ obtains the same results with it on eight functions ($f_1$-$f_5$, $f_{10}$, $f_{16}$, $f_{18}$). Besides, their performances are very close to each other on three functions ($f_6$, $f_{15}$, $f_{19}$). But C-EDA$^2$ is surpassed by RS-CMSA on the remaining functions.

In summary, C-EDA$^2$ indicates better performance than LMCEDA and LMSEDA, which is achieved with much simple algorithmic framework and less parameters. But C-EDA$^2$ also has some shortages, there is still room to improve its performance. The major disadvantage of C-EDA$^2$ lies in the restart mechanism. Independent restarts may improve its performance to some extent, it is still very likely to revisit previously explored regions. While in RS-CMSA, taboo method is adopted to reduce the chance of revisiting in restarts, which makes RS-CMSA more efficient. Similar idea could also be introduced to C-EDA$^2$ to further enhance its performance.

Table 3: Average peak ratios obtained by C-EDA$^2$, LMCEDA, LMSEDA and RS-CMSA on the 20 functions of CEC'2013 multimodal function set at accuracy level $10^{-4}$.

| Fun. | LMCEDA | LMSEDA | RS-CMSA | C-EDA$^2$ |
| --- | --- | --- | --- | --- |
| $f_1$ | 1.000 | 1.000 | 1.000 | 1.000 |
| $f_2$ | 1.000 | 1.000 | 1.000 | 1.000 |
| $f_3$ | 1.000 | 1.000 | 1.000 | 1.000 |
| $f_4$ | 1.000 | 1.000 | 1.000 | 1.000 |
| $f_5$ | 1.000 | 1.000 | 1.000 | 1.000 |
| $f_6$ | 0.990 | 0.972 | 0.999 | 1.000 |
| $f_7$ | 0.734 | 0.673 | 0.997 | 0.711 |
| $f_8$ | 0.347 | 0.613 | 0.871 | 0.839 |
| $f_9$ | 0.284 | 0.248 | 0.730 | 0.300 |
| $f_{10}$ | 1.000 | 0.998 | 1.000 | 1.000 |
| $f_{11}$ | 0.667 | 0.892 | 0.997 | 0.667 |
| $f_{12}$ | 0.750 | 0.990 | 0.948 | 0.663 |
| $f_{13}$ | 0.667 | 0.667 | 0.997 | 0.667 |
| $f_{14}$ | 0.667 | 0.667 | 0.810 | 0.667 |
| $f_{15}$ | 0.696 | 0.738 | 0.748 | 0.740 |
| $f_{16}$ | 0.667 | 0.670 | 0.667 | 0.667 |
| $f_{17}$ | 0.456 | 0.620 | 0.703 | 0.660 |
| $f_{18}$ | 0.657 | 0.660 | 0.667 | 0.667 |
| $f_{19}$ | 0.451 | 0.458 | 0.503 | 0.500 |
| $f_{20}$ | 0.059 | 0.248 | 0.483 | 0.248 |

## 5 CONCLUSIONS

This paper first introduces a novel archive-based EDA named EDA$^2$ and further proposes a niching EDA called C-EDA$^2$ by combing EDA$^2$ with an adaptive clustering strategy. Different from most existing EDAs which only employ some good solutions in the current population to build their probability models, EDA$^2$ maintains an archive to preserve the high-quality solutions generated in a certain number of previous generations and uses these solutions to assist estimating the Gaussian model. This simple operation naturally integrates the evolution direction information into the estimated covariance matrix, which significantly improves the search efficiency of EDA$^2$. Moreover, the population size of EDA$^2$ is also greatly reduced by utilizing historical solutions. Experiments on typical test functions demonstrate that EDA$^2$ is simple and efficient, it is also very robust to its parameters. Then EDA$^2$ is further incorporated into an adaptive clustering strategy for solving multimodal optimization problems. Comparison results on a set of niching benchmark functions indicate that C-EDA$^2$ is competitive to two well-established multimodal EDAs.

However, even though C-EDA$^2$ shows its potential in dealing with multimodal problems, there is still room to further improve its performance. For future research, the restart mechanism in C-EDA$^2$ could be improved by combining techniques like taboo method. Besides, it is also interesting to equip EDA$^2$ with other niching schemes.

## ACKNOWLEDGMENTS
This work was supported in part by the National Natural Science Foundation of China under Grants 61105126 and in part by the Postdoctoral Science Foundation of China under Grants 2014M560784 and 2016T90922.